\documentclass[runningheads]{llncs}
\usepackage{graphicx}
\usepackage{url}
\usepackage{color, graphicx}

\usepackage[compatibility=false]{caption}
\usepackage{subcaption}

\usepackage{multirow}

\newtheorem{mymethod}{Method}
\newtheorem{mydefinition}{Definition}
\newtheorem{myassumption}{Assumption}
\newtheorem{myobservation}{Observation}
\newtheorem{myremark}{Remark}
\newtheorem{myproposition}{Proposition}
\newtheorem{myclaim}{Claim}
\newtheorem{mylemma}{Lemma}
\newtheorem{mycorollary}{Corollary}
\newtheorem{myexample}{Example}
\newtheorem{myexamples}{Examples}
\newtheorem{myalgorithm}{Algorithm}
\newtheorem{myconstruction}{Construction}
\newtheorem{myrule}{Rule}

\newcommand{\bolddot}{\hspace{-1.5mm}\textbf{.}\ \  }

\newcommand{\BT}{\begin{theorem}}
\newcommand{\ET}{\end{theorem}}
\newcommand{\BCR}{\begin{mycorollary}\bolddot}
\newcommand{\ECR}{\end{mycorollary}}
\newcommand{\BAS}{\begin{myassumption}}
\newcommand{\EAS}{\end{myassumption}}
\newcommand{\BPR}{\begin{myproposition}}
\newcommand{\EPR}{\end{myproposition}}
\newcommand{\BL}{\begin{mylemma}\bolddot}
\newcommand{\EL}{\end{mylemma}}
\newcommand{\BCM}{\begin{myclaim}\bolddot}
\newcommand{\ECM}{\end{myclaim}}

\newcommand{\BD}{\begin{mydefinition}}
\newcommand{\ED}{\end{mydefinition}}
\newcommand{\BPF}{\begin{proof}}
\newcommand{\EPF}{\end{proof}}
\newcommand{\BEX}{\begin{myexample}}
\newcommand{\EEX}{\end{myexample}}
\newcommand{\BEXS}{\begin{myexamples}}
\newcommand{\EEXS}{\end{myexamples}}
\newcommand{\BOB}{\begin{myobservation}}
\newcommand{\EOB}{\end{myobservation}}
\newcommand{\BR}{\begin{myremark}}
\newcommand{\ER}{\end{myremark}}
\newcommand{\BAL}{\begin{myalgorithm}}
\newcommand{\EAL}{\end{myalgorithm}}
\newcommand{\BAM}{\begin{mymethod}}
\newcommand{\EAM}{\end{mymethod}}

\newcommand{\BCO}{\begin{myconstruction}}
\newcommand{\ECO}{\end{myconstruction}}
\newcommand{\BRule}{\begin{myrule}}
\newcommand{\ERule}{\end{myrule}}

\newcommand{\BE}{\begin{enumerate}}
\newcommand{\EE}{\end{enumerate}}
\newcommand{\BI}{\begin{itemize}}
\newcommand{\EI}{\end{itemize}}

\newenvironment{Rightitem}{%
  \begin{itemize}}{\end{itemize}}
\newenvironment{Leftitem}{%
    \begin{itemize}}{\end{itemize}}
\newenvironment{Iffitem}{%
    \begin{itemize}}{\end{itemize}}
\newenvironment{Iffitemi}{%
    \begin{itemize}}{\end{itemize}}

\newcommand{\BRI}{\begin{Rightitem}\item}  \newcommand{\ERI}{\end{Rightitem}}
\newcommand{\BLI}{\begin{Leftitem}\item}   \newcommand{\ELI}{\end{Leftitem}}
\newcommand{\BIFF}{\begin{Iffitem}\item}   \newcommand{\EIFF}{\end{Iffitem}}
\newcommand{\BIFFI}{\begin{Iffitemi}\item}   \newcommand{\EIFFI}{\end{Iffitemi}}

\newcommand{\bc}{\begin{center}}
\newcommand {\ec}{\end{center}}

\newcommand{\stam}[1]{}

\newcommand{\bd}{\begin{definition}}
\newcommand{\ed}{\end{definition}}
\newcommand{\be}{\begin{enumerate}}
\newcommand{\bi}{\begin{itemize}}
\newcommand{\ee}{\end{enumerate}}
\newcommand{\ei}{\end{itemize}}

\begin{document}

\title{Towards a Framework to Manage \protect\\ Perceptual Uncertainty for \protect\\ Safe Automated Driving}

\titlerunning{Managing Perceptual Uncertainty for Safe Automated Driving}

\author{Krzysztof Czarnecki \and
Rick Salay}

%

\institute{University of Waterloo, Waterloo, Canada \\
\email{\{kczarnec, rsalay\}@gsd.uwaterloo.ca}}

\maketitle              
\begin{abstract}
Perception is a safety-critical function of autonomous vehicles and machine learning (ML) plays a key role 
in its implementation.  This position paper identifies (1) perceptual uncertainty as a performance measure used to define safety requirements
and (2) its influence factors when using supervised ML.
This work is a first step towards a framework for measuring and controling the effects of these factors 
and supplying evidence to support claims about perceptual uncertainty. 

\end{abstract}
\section{Introduction}\label{sec:intro}
An Autonomous Driving System (ADS) consists of components for perception of the environment and vehicle state, planning the vehicle actions, and controlling the vehicle to implement these actions. While ML can play a role in all three areas, in the position paper we focus attention on perception and the use of supervised ML to infer the state information. Furthermore, we take an uncertainty-centric view of safety.
The level of assurance in the safety of an ADS could be expressed in terms of our uncertainty that it will behave acceptably safely in all relevant situations. Then our assurance objective is to reduce this uncertainty to an acceptable level. This is different, but complementary, to the system safety objective of reducing risk to an acceptable level.


Uncertainty is an established measure of performance when perceiving the world. A desired property of perception is \emph{accuracy}, that is, the estimated state being close to the true state of the world. 
In engineering measurement practice, the concept of accuracy has been largely abandoned and replaced by uncertainty estimation, however. For example, the ``Guide to the expression of uncertainty in measurement'' (GUM)~\cite{GUM} considers accuracy as a qualitative concept since the true value of the measured physical quantity is generally unknown and unknowable. On the other hand, measurement uncertainty, defined as a characterization of the dispersion of the values that could reasonably be attributed to the quantity being measured~\cite{GUM}, can be estimated based on observable quantities and available knowledge. Note that measurement errors due to systematic effects (i.e., measurement bias) can be reduced if the effect is identified and a correction is applied. However, uncertainty will remain about whether the
correction adequate. The GUM requires that sources of uncertainty associated with both random and systematic effects in the measurement procedure, instruments, and the particular measurement are rigorously identified and assessed.

In this position paper, we propose that this recommendation should apply to \emph{all perception tasks}, not just the measurement of physical quantities. Perception can be seen as a form of measurement of state and has an associated \emph{perceptual uncertainty}. For example, classifier accuracy on a test set is one observable quantity that can be used to estimate the overall uncertainty of a classifier in the field, but additional influence factors, such as the test set quality and representativeness, must also be considered. When a perceptual component is safety-critical, identifying the factors that influence perceptual uncertainty is a key step in assuring safety.

Our aim in this paper is to outline a generic framework for managing perceptual uncertainty in order to support the systematic safety assurance of perception components. Specifically, our contributions are the following: (1) we present a generic model for thinking about perception components called the \emph{perception triangle}; (2) we discuss how perceptual uncertainty can be seen as a performance measure used to define safety requirements; and, (3) we identify as set of factors that influence perceptual uncertainty and discuss their measurement and impacts.


\section{Perception and Uncertainty} \label{sec:perunc}

\begin{figure} [h]
	\centering{
		\vspace{-0.2in}
		\begin{subfigure}[b]{0.39\textwidth}
			\includegraphics[width=\textwidth]{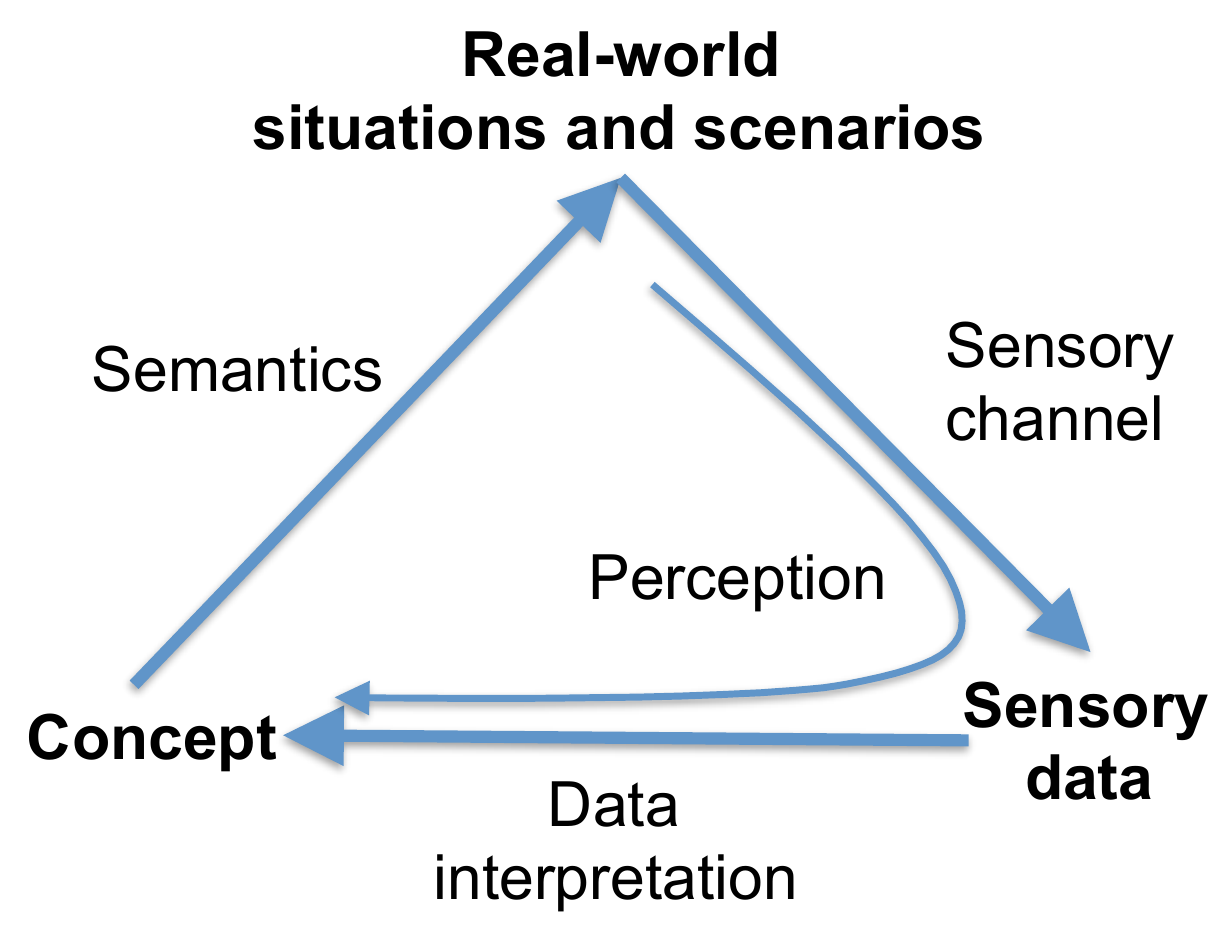}
			\caption{Domain level \protect\\ (generic)}
			\label{fig:gpertri}
		\end{subfigure}
		\begin{subfigure}[b]{0.51\textwidth}
			\includegraphics[width=\textwidth]{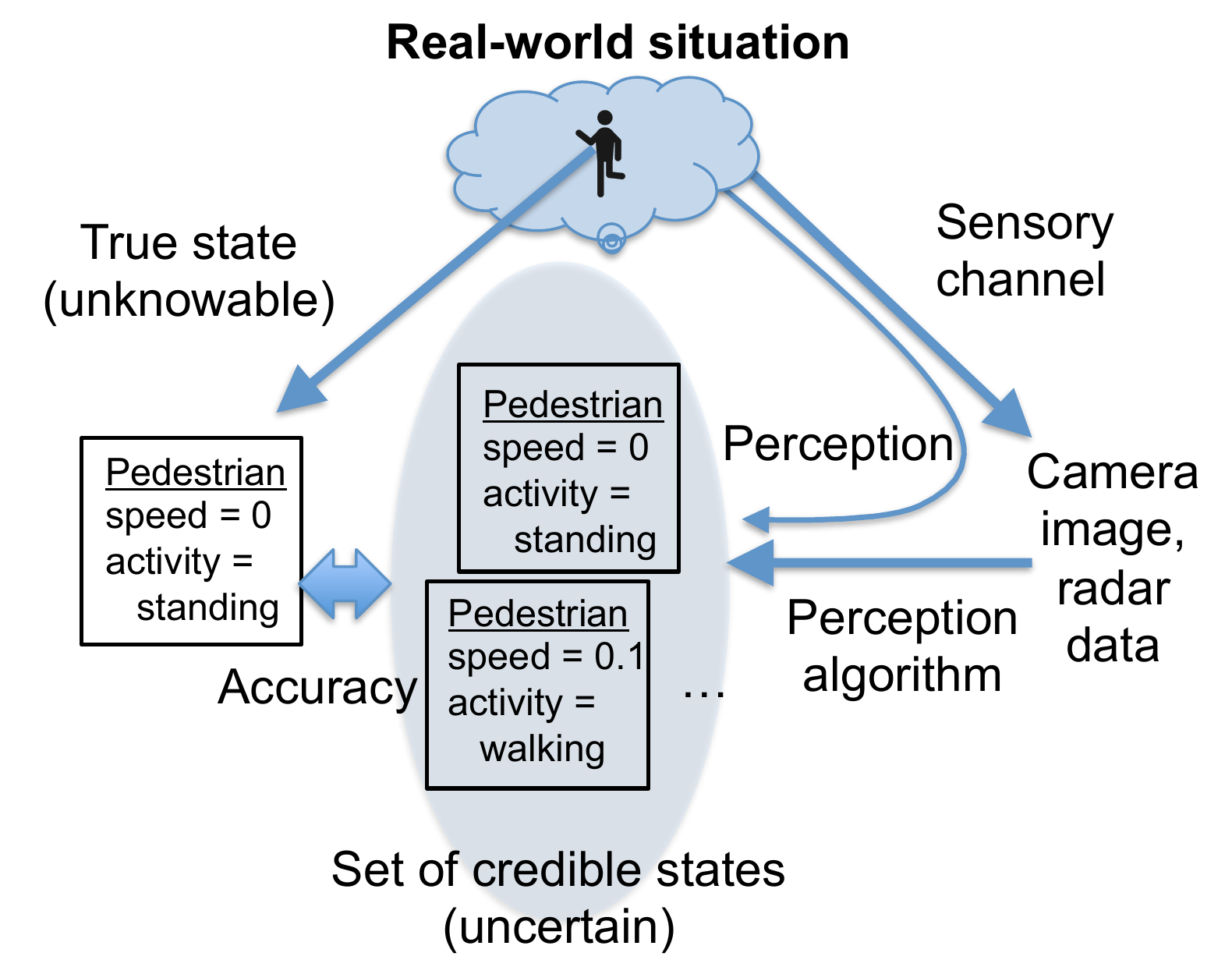}
			\caption{Instance level example \protect\\ (pedestrian concept \& specific situation)}
			\label{fig:ipertri}
	\end{subfigure}}
	\caption{Perception triangle}
	\vspace{-0.25in}
	\label{fig:pertri}
\end{figure}

An ADS relies on perception to discover the current state of the subject vehicle and its driving environment and track the state over time.
Figure~\ref{fig:gpertri} illustrates the perception task using a \emph{perception triangle}, which is loosely inspired by the semiotic triangle. 

The objective of perception is to create a conceptual representation of the real world that captures facts relevant to the task.
A concept defines the structure of the representation, including relevant attributes and relations.
For example, a pedestrian is a person within a road who is neither a vehicle occupant nor a rider.
Pedestrian attributes relevant to driving include pedestrian pose and extent, dynamic state, and activity.
A mapping from the concept to real-world situations and scenarios identifies concept instances in the situations and scenarios, and thereby defines the concept semantics.
Perception, indicated by the bent arrow, achieves the inverse task of linking real-world situations to their conceptual representation in two steps.
First the sensory channel produces sensory data, such as images and radar returns.
Then perception algorithms interpret the data to create the conceptual representation of the real world.
The sensory channel represents the information pathway from the objects in the real-world scene to one or more sensors.
Figure~\ref{fig:gpertri} represents perception at the domain level, which includes all conceivable situations and scenarios to define the concept of interest.
The next section gives an example of an instance-level realization of the triangle for perceiving a particular real-world situation or scenario.

\subsection{Perceptual Uncertainty as a Performance Measure}

Perception yields an uncertain, or limited knowledge of the true state of the world, which is called \emph{perceptual uncertainty}.  Figure~\ref{fig:ipertri} illustrates this idea by instantiating the perception triangle for the pedestrian concept and a particular real-world situation. The situation contains a pedestrian who is visible to the camera and radar sensors of the subject vehicle. The true state of the pedestrian may be ``standing still'', but the perception algorithm interprets the sensory data and comes up with its own hypotheses about the state (see the ellipse in Figure~\ref{fig:ipertri}). Probability mass or density function over the possible world states can model this uncertainty, where probability represents the credibility of any of the states being the true state given the knowledge of the overall perception setup. The perception algorithm may compute either an estimate of this probability distribution over the states or a point estimate of the state. In the latter case, one may use existing knowledge about the perception system, such as test results, to reason about the uncertainty of the point estimate.

The goal of ADS perception is to establish facts about the real world with acceptable perceptual uncertainty and sufficient timeliness as required for the dynamic driving task. The required perception performance depends on the type of fact to be established, the driving situation, and the overall ADS performance to be achieved. 
For example, perception of minor roughness is not safety relevant, but may be required for driving comfort;
however, perception of major pavement damage, such as big potholes, is safety relevant.

ADS safety is the absence of unreasonable risk of crashes due to inadequate behavior, malfunctions, or security vulnerabilities. Existing practices and standards address safety assurance related to software and hardware malfunctions~\cite{ISO26262} and security vulnerabilities~\cite{sae2016j3061}. Our focus is on unreasonable risk caused by inadequate perception performance. The safety requirements on perception are established through hazard analysis and risk assessment, which considers different situations and scenarios that the system can encounter in operation, as defined by its Operational Design Domain (ODD)~\cite{sae2013j3016}. The resulting safety requirements specify what types of facts need to be perceived and the bounds on the uncertainty in different driving situations in order to keep the crash risk reasonable. Thus the safety requirements analysis must consider the crash risk caused by an incorrect perceptual decision and the reasonableness of this risk. 


Figure~\ref{fig:gpertri} illustrates the key elements of safety requirements on perception. The concept specification should list the states to be perceived.
For pedestrian detection, they may include pedestrian pose and extent, dynamic state, and activity; however, the specification should also provide \emph{additional 
concept features} that may impact pedestrian perception in the context of the ODD, such as different illumination and occlusion levels, clothing variations, pedestrian traffic density, and situation context. The performance specification on perception (bent arrow) should provide scenario-dependent detection range and bounds on detection uncertainty and timing.

\subsection{Factors Influencing Perceptual Uncertainty When Using Machine Learning}

\begin{figure} [h]
	\centering{
		\vspace{-0.3in}
		\includegraphics[width=.7\textwidth]{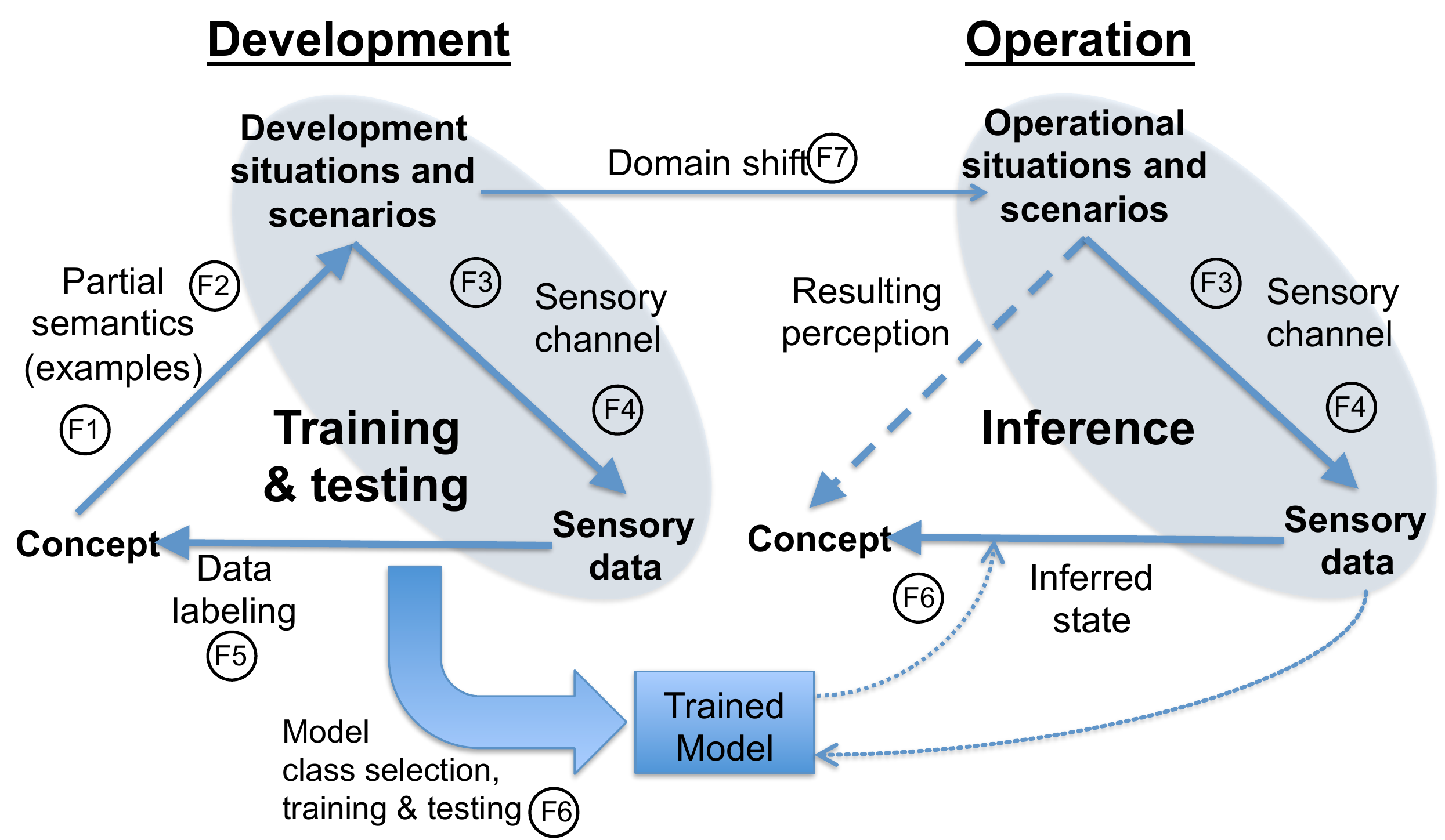}}
	\caption{Sources of perceptual uncertainty when using machine learning.}
	\vspace{-0.2in}
	\label{fig:usources}
\end{figure}

In order to meet the specification, the sources of perceptual uncertainty must be identified and controlled. Figure~\ref{fig:usources} implements the specification-level perceptual triangle from Figure~\ref{fig:gpertri} using supervised ML. The figure shows two implementation-level perceptual triangles---one for development and another for operation (we assume no learning in operation). The left triangle illustrates training and testing during development.  A necessarily partial set of situations and scenarios exemplifies the concept semantics for training and testing during development. The recorded sensory data is typically passed to human labelers, who interpret and label it. The training and testing process (shown as the fat arrow) uses the resulting labeled data to produce a trained model. The right triangle illustrates inference during operation. The trained model interprets the sensory data from the situations and scenarios in operation and produces the estimated state. The dashed arrow summarizes the overall perception as the composition of sensing and inference.

Figure~\ref{fig:usources} identifies the factors that influence the overall perceptual performance in operation. Their influence and interdependence are as follows.

\emph{Conceptual uncertainty} (F1): 
The concept definition may allow different interpretations by different stakeholders. Conceptual uncertainty influences the selection of development situations and scenarios (F2) by making developers uncertain whether a particular situation or scenario is relevant to exemplify the concept. It also increases labeling uncertainty (F5) by increasing the chances that different labelers will interpret the concept differently. Conceptual uncertainty can be assessed qualitatively, such as by an expert review, or quantitatively, such as through labeling disagreement statistics and additional labeler feedback. The latter requires an enhanced labeling effort that also includes the additional concept and scenario features relevant to F2 and F3.

\emph{Development situation and scenario coverage} (F2): This factor is the degree to which the situations and scenarios used in training and testing cover the specified concept and ODD. It impacts model uncertainty (F6) and, through it, the perceptual uncertainty, which will be reduced if instances of situations and scenarios in specification scope are missed or are too few. Enhanced labeling of training and test datasets with additional concept and scenario features allows computing coverage and frequency statistics to guide the subsequent improvement of these sets; however, extensive validation testing and data collection in the field are necessary to discover so-called \emph{unknown unknowns}, i.e., new relevant situations and scenarios that cannot be constructed from knowledge collected so far~\cite{koopmantoward}.  

\emph{Situation or scenario uncertainty} (F3): The structure of a situation or scenario may limit the observability of the state of interest and thereby allow multiple interpretations. As a result, the perceptual uncertainty will increase. This factor can be assessed through measures such as distance, levels of occlusion, illumination, and clutter, which were already mentioned under F1 and F2. For example, pedestrian distance and occlusions will limit the amount of information about the pedestrian that reaches the vehicle sensors. The impact of these measures on perceptual uncertainty should be assessed in testing.

\emph{Sensor properties} (F4): Sensor properties, in combination with the situation and scenario, may limit the amount of information of interest that is sensed, and thereby increase the perceptual uncertainty. Important sensor properties include sensing mode, range, resolution, noise characteristics, calibration, and placement. The sensor requirements are determined from perception requirements using established methods from sensor engineering~\cite{bussemaker2014sensing}.
The impact of sensor properties on the overall perception result, especially given a trained model, should be assessed through sensitivity studies.

\emph{Labeling uncertainty} (F5): Human labelers may disagree due to limited knowledge of the concept (F1) or the uncertainty from F3-4 or simply make accidental mistakes when labeling. Established methods exist to measure and reduce this uncertainty~\cite{hung2013evaluation}.

\emph{Model uncertainty} (F6): There is an uncertainty of what the model has learned in training and what decisions it will make in operation. This uncertainty depends on model class, capacity, and training data and procedure. Confidence measures (e.g., Bayesian deep learning~\cite{gal2016dropout}) can be used to provide a run-time measure of model uncertainty. For example, a low confidence pedestrian detection indicates a potential weakness in the model which must be diagnosed and addressed (e.g., by adding training data, changing the model class, etc.)

\emph{Operational domain uncertainty} (F7): There is also uncertainty whether the situations and scenarios encountered in operation were adequately reflected in training; and whether the sensor properties in operation match those in training. For example, sensors in operation may be misaligned or calibrated differently. Novelty detection methods have been proposed to detect out-of-distribution inputs (i.e., outliers). For example, the training dataset can be used without labels to train an autoencoder to identify the characteristics of the dataset. At run-time, new inputs are checked for their similarity to the dataset (i.e., their novelty) by checking how well the autoencoder can reconstruct them~\cite{sabokrou2018adversarially}. The frequency of novel inputs encountered at run-time is a measure of operational domain uncertainty (F7).



%
%
%

\section{Conclusion and Next Steps}\label{sec:discuss}
In this position paper, we present first steps toward the development of a generic framework for perception safety that focuses on reducing perceptual uncertainty to an acceptable level. We identified a set of factors that contribute to perceptual uncertainty and briefly discussed their measurement and impact. Several topics are logical next steps. First, the measurement of the factors and their effects on perceptual uncertainty requires a systematic analysis; in addition to measurement at development time, F3,4,6,7 should also be assessed in operation. Next, methods to control, that is, eliminate or reduce, the negative effects of these factors on perceptual uncertainty, and if not possible, to mitigate the effects, need to be provided. Finally, the argument structures and types of evidence to support claims about perceptual uncertainty in safety cases need to be established.


\bibliographystyle{splncs04}
\bibliography{references}

\end{document}